\documentclass[10pt,journal,compsoc]{IEEEtran}
\usepackage{caption2}

\ifCLASSOPTIONcompsoc
  \usepackage[nocompress]{cite}
\else
  \usepackage{cite}
\fi
\ifCLASSINFOpdf
   \usepackage[pdftex]{graphicx}
  \graphicspath{{../pdf/}{../jpeg/}}
  \DeclareGraphicsExtensions{.pdf,.jpeg,.png}
\else
  \usepackage[dvips]{graphicx}
  \graphicspath{{../eps/}}
  \DeclareGraphicsExtensions{.eps}
\fi
\usepackage{amsmath}
\begin{document}
%
\title{Uncover Common Facial Expressions in Terracotta Warriors: A Deep Learning Approach}
%
%
%
%

\author{Wenhong Tian*,
        Yuanlun Xie,
        Tingsong Ma, Hengxin Zhang
        
\IEEEcompsocitemizethanks{\IEEEcompsocthanksitem Wenhong Tian is with School of Information and Software Engineering, University of Electronic Science and Technology of China\protect\\
E-mail: tian\_wenhong@uestc.edu.cn}}
\IEEEtitleabstractindextext{%
\begin{abstract}
Can advanced deep learning technologies be applied to analyze some ancient humanistic arts? Can deep learning technologies be directly applied to special scenes such as facial expression analysis of Terracotta Warriors? The big challenging is that the facial features of the Terracotta Warriors are very different from today's people. We found that it is very poor to directly use the models that have been trained on other classic facial expression datasets to analyze the facial expressions of the Terracotta Warriors. At the same time, the lack of public high-quality facial expression data of the Terracotta Warriors also limits the use of deep learning technologies. Therefore, we firstly use Generative Adversarial Networks (GANs) to generate enough high-quality facial expression data for subsequent training and recognition. We also verify the effectiveness of this approach. For the first time, this paper uses deep learning technologies to find common facial expressions of general and postured Terracotta Warriors. These results will provide an updated technical means for the research of art of the Terracotta Warriors and shine lights on the research of other ancient arts. 
\end{abstract}

}

\maketitle

\IEEEdisplaynontitleabstractindextext

\IEEEpeerreviewmaketitle

\IEEEraisesectionheading{\section{Introduction}\label{sec:introduction}}

\IEEEPARstart{F}{acial} expression has a significant impact on human daily communication. Automatically recognizing facial expressions can help computers understand human expressions and realize human-computer interaction. During recent years, facial expression recognition (FER) has become a hot topic in the field of computer vision, which aims to recognize human’s expression categories through facial images. As a result, many deep learning based facial expression recognition methods have been proposed~\cite{1,2,3,4,5,6,7,8,9,10}.

\begin{figure*}[h]
\centering
\includegraphics[width=0.95\textwidth]{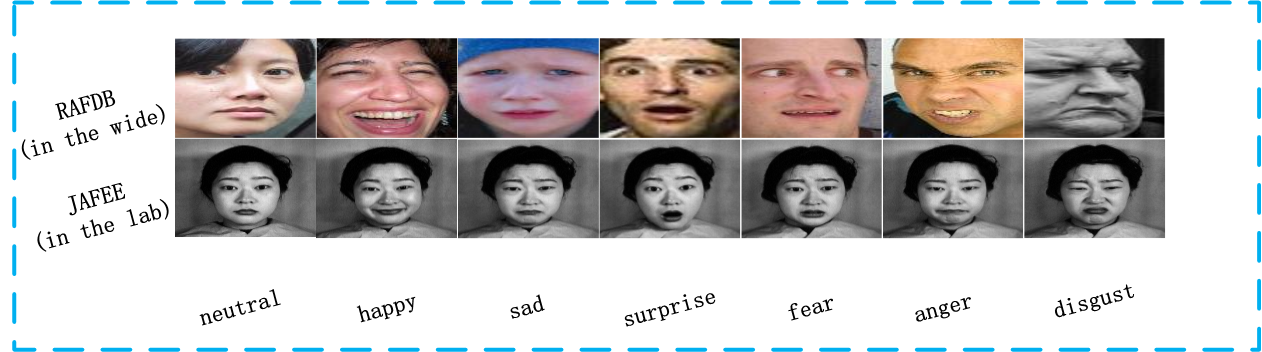}
\caption{The facial expression samples from two datasets.}
\label{fig:dataset}
\end{figure*}

~~Regularly, facial expression recognition mainly has two types of environments: in the lab and in the wild. In the early stage, researchers always use face images that are controlled and constrained in the lab, for instance, with a fixed camera angle or a specific light intensity. There are some well-known datasets for the lab environment, such as JAFFE~\cite{10}, MMI~\cite{11} and CK+~\cite{12}. As for the wild environment, other well-known datasets, for instance, FERPlus ~\cite{13}, RAFDB~\cite{14}, AffectNet~\cite{15} are proposed. Some samples can be seen in  Figure \ref{fig:dataset}. In summary, good progress has been made in the lab environment~\cite{8}, however, the research has made slow progress in the wild environment, because the images in the wild are obtained in an unconstrained way, such as with weak light conditions and fuzzy images and so on. \\

There are two phases in facial expression recognition: feature extraction and feature classification. In the feature extraction phase, the network mainly extracts the low-level and high-level feature representation in the face image for different facial expressions. The quality of the extracted facial expression features has a vital influence on the accuracy and generalization ability of facial expression recognition. In the feature classification phase, the network classifies the extracted features, and provides the probability of each category.\\

Recently, most researchers usually use advanced feature extraction methods represented by deep convolutional neural network~(CNN) models to improve the accuracy of facial expression recognition in the wild. Li et al.~\cite{16} propose a Deep Bi-Manifold CNN to learn the discriminative feature for multi-label expressions, the experimental results show that their method has good recognition performance for most facial expression datasets. Slimani et al.~\cite{17}, for the first time, introduce a Highway Convolutional Neural Network~(HCNN) to the facial expression recognition field, and achieved a test accuracy of 52.14\% on CFEE dataset~\cite{15}. Li et al.~\cite{18} proposes a deep learning method with attention mechanism, which can automatically perceive the occluded area of the face, and focuses on the unoccluded area that can provide more effective facial expression features. In order to solve the pose and occlusion problems in facial expression recognition, Wang et al.~\cite{19} proposes a network that is based on regional attention, after evaluating on multiple datasets, their method improves the performance on FER. Wang et al. ~\cite{6} propose a Self-Cure Network~(SCN) to suppresses the uncertainties efficiently and prevents deep networks from over-fitting uncertain facial images~(ambiguous facial expressions, low-quality facial images,~and the subjectiveness of annotators), and achieves competitive accuracy of 88.14\% on RAF-DB, 60.23\% on AffectNet, and 89.35\% on FERPlus respectively against the-state-of-the-art results. Shi et al.~\cite{20} proposes a novel Amend Representation Module (ARM) on ResNET18~\cite{26} to reduce the weight of eroded features to offset the side effect of padding, and share affinity features over minibatch to strengthen the representation learning, they achieved average accuracy respectively 90.55\% on RAF-DB, 64.49\% on Affect-Net, and 71.38\% on FER2013, exceeding current state-of-the art methods; they provided source codes online so that one can easily reproduce the method.\

Although many good methods have been introduced, they have not achieved ideal performance in the wild. One reason is that many low-level feature information is lost and the extracted features only contain high-level abstract features during feature extraction phase. Similarly, in the classification phase, existing methods rarely consider how to better fuse features of different levels to improve the accuracy. Current authors~\cite{21} proposed an approach to combing extraction and fusion of different levels of features.  The recognition accuracy is also affected by factors such as age and skin color that are derived from the facial image itself, so that it is still a big challenge to recognize with high accuracy in the wild.\\
Cowen et al.~\cite{22} introduced a traditional computational approach to find universal facial expressions in art of the ancient Americas, where Pain,Strain,Anger, Elation and Sadness are five universal facial expressions they found. Cowen et al.~\cite{23} found sixteen facial expressions preserved across the modern world using machine-learning approach.  Lillicrap et al.~\cite{28} present a surprisingly simple mechanism that assigns blame by multiplying errors by even random synaptic weights for deep learning. Wei et al.~\cite{29} apply deep learning to accurately classify primary and metastatic cancers. \\


The research on the facial expressions of general and postured Terracotta Warriors is helpful to analyze their cultural attributes and characteristics. It can also be applied to artistic creation, including the production of expression bags. Although some studies have used traditional methods to recognize and classify the expressions of the Terracotta Warriors, there is still lack of research on the recognition of the seven basic expressions (Neutral, Happy, Anger, Sad, Surprise, Fear, Disgust) of the Terracotta Warriors by using deep learning technology in the published literature. \\

There are two major challenges for the study of the facial expressions of Terracotta Warriors based on deep learning methods: firstly, the lack of datasets puts barrier to this task, and secondly, the facial expressions of Terracotta Warriors have a clear gap from real-human facial expressions, leading to a certain difficulty in the application of deep learning technology. We find that the deep learning models pre-trained on real-human facial expressions cannot be used to directly analyze the facial expressions of Terracotta Warriors, the models need more Terracotta Warriors datasets to be trained from scratch to obtain the true facial distribution. To address these two challenges, we firstly apply a GAN model with great performance to generate more Terracotta Warriors images to solve the problem of lack of Terracotta Warriors datasets, then we use well trained model to uncover common facial expressions of general and postured Terracotta Warriors. The five considered postures are kneeling soldiers, Generals, vertical soldiers, cavalries, civil servants. \\
In summary, the main contributions of our work include: \\
 (1) For the first time, we propose a deep learning approach to uncover common facial expressions in general and postured Terracotta Warriors.
 
 (2) To address the problem of lack of dataset, we firstly apply GANs to generate enough high-quality photos for Terracotta Warriors.
 
 (3) We conduct extensive experiments including training deep learning models to uncover common facial expressions in general and postured Terracotta Warriors, these approaches also shine lights on the research of other ancient arts.
 
%



 
 

%

\textbf{\begin{table*}[h]
\centering
 \caption{ \label{fig:RAFDB trainset result} The performance of the ARM on the RAFDB trainset.}
 \begin{tabular}[t]{lllllllll}
 \hline
 & Surprise & Fear & Disgust & Happy & Sad & Anger & Neutral & Average \\ \hline
 Accuracy & 1.0 & 0.999 & 1.0 & 1.0 & 1.0 & 1.0 & 1.0 & 0.999\\ \hline
 \end{tabular}
\end{table*}}

\textbf{\begin{table*}[h]
\centering
 \caption{ \label{fig:RAFDB testset result} The performance of the ARM on the RAFDB testset.}
 \begin{tabular}[t]{lllllllll}
 \hline
 & Surprise & Fear & Disgust & Happy & Sad & Anger & Neutral & Average \\ \hline
 Accuracy & 0.86 & 0.703 & 0.681 & 0.952 & 0.866 & 0.741 & 0.968 & 0.901 \\ \hline
 \end{tabular}
\end{table*}}

\begin{table*}[h]
\centering
 \caption{ \label{fig:TW58 test accuracy} The performance of the ARM on TW58 dataset(pre-trained on RAFDB).}
 \begin{tabular}[t]{lllllllll}
 \hline
 & Happy & Sad & Anger & Neutral & Fear & Surprise & Disgust & Average \\ \hline
 Accuracy & 0.588 & 1.0 & 0.0667 & 0.0 & 0.0  & 0.0 & 1.0 & 0.216 \\ \hline
 \end{tabular}
\end{table*}


 

\section{Results}
In this section, we mainly conduct experiments on three datasets: TW58 dataset of Terracotta Warriors of Qin and Han Dynasty, GAN18K (a total number of 18690 pictures generated by GAN) and dataset of Terracotta Warriors with different postures. Then we present the implementation details of the experiments. We manually annotate the collected dataset of Terracotta Warriors online and from the digital library of Qin Terracotta Warriors, the total number of usable images is 58 (denoted as TW58). The labeling was carried out by six team members independently, it is worth noting that the expression images of Terracotta Warriors are difficult for human eyes to distinguish, and the same expression image may be considered as multiple expressions (generally 1 to 3 expression categories). The majority of the original labels is adopted as the final label for each image. Based on TW58, we generate 18690 different expressions using GAN (called GAN18K). It is worth noting that we use ARM~(20) model in our whole training and testing process with comparison to ResNeT18(16).

\subsection{Common Facial Expressions in Terracotta Warriors}
\subsubsection{Testing on TW58 dataset}
Firstly, we adopt the deep learning model to train on the RAFDB dataset, and obtain the pre-trained model to test the TW58 dataset. As we can see that in Table \ref{fig:RAFDB trainset result} and Table \ref{fig:RAFDB testset result}, the deep learning model performs well in the RAFDB trainset and testset respectively, so we can use the pre-trained model to test the high-quality Terracotta Warriors images. Table \ref{fig:TW58 test accuracy} shows experimental test results on the TW58 dataset using pre-trained model on RAFDB, we can observe that the recognition accuracy of Disgust,Sad and Happy is much higher than other expressions, which is respectively 100\%, 100\% and 58.8\% while the average recognition accuracy is 21.6\%.

\subsubsection{Evaluation on images generated by GAN}
In this section, we use the images generated by GAN method to train from scratch and test. According to the normal proportion of trainset and testset in deep learning, we divide the dataset generated by GAN roughly into (train set): (validate set): (test set) = 7:1:2. In this GAN-generated dataset (called GAN18K dataset), 15730 pictures are used for training and 2960 pictures are for testing. First of all, we use the ResNet-18~\cite{26} network to carry out the experimental evaluation, which has the best performance in the current deep learning field, however, we find the test results are not ideal, such as those in the first row of the Table \ref{fig:GAN train accuracy}, Table \ref{fig:GAN validation accuracy} and Table \ref{fig:GAN test accuracy}. Then, we use ARM network to carry out the experimental evaluation, which is one of the best methods reported in facial expression recognition. The second row of Table \ref{fig:GAN train accuracy}, Table \ref{fig:GAN validation accuracy} show our experimental results on the images generated by GAN; in Table \ref{fig:GAN test accuracy} where ARM-1 results are obtained by strictly partitioning the  dataset so that there is no image (generated from same input figure) appear both training and test dataset, ARM-2 results are obtained by partitioning the  dataset so that 70\% images generated from same input figure are in training dataset and the remaining 30\% images  are in test datset.We can observe that the recognition accuracy of Surprise, Anger and Happy is much higher than other expressions, which is respectively 97.58\%, 82.31\%, and 91.67\% from ARM-1 while respectively 94\%, 95\%, and 96.4\% from ARM-2.

\textbf{\begin{table*}[h]
\centering
 \caption{ \label{fig:GAN train accuracy} The train-set accuracy of the ResNet-18 and ARM on images generated by GAN.}
 \begin{tabular}[t]{lllllllll}
 \hline
 & Surprise & Fear & Disgust & Happy & Sad & Anger & Neutral & Average \\ \hline
 ResNet-18 & 1.00 & 0.44 & 1.00 & 1.00 & 0.85 & 0.99 & 0.99 & 0.90 \\ \hline
 ARM & 1.00 & 1.00 & 1.00 & 1.00 & 0.998 & 1.00 & 1.00 & 0.999 \\ \hline
 \end{tabular}
\end{table*}}

\textbf{\begin{table*}[h]
\centering
 \caption{ \label{fig:GAN validation accuracy} The validate-set accuracy of the ResNet-18 and ARM on images generated by GAN.}
 \begin{tabular}[t]{lllllllll}
 \hline
 & Surprise & Fear & Disgust & Happy & Sad & Anger & Neutral &Average\\ \hline
 ResNet-18 & 0.70 & 0.29 & 0.54 & 0.55 & 0.28 & 0.53 & 0.30 &0.44 \\ \hline
 ARM & 0.95 & 0.32 & 0.80 & 0.89 & 0.617 & 0.851 & 0.672 &0.685 \\ \hline
 \end{tabular}
\end{table*}}
\textbf{\begin{table*}[h]
\centering
 \caption{ \label{fig:GAN test accuracy} The test-set accuracy of the ResNet-18 and ARM on images generated by GAN.}
 \begin{tabular}[t]{lllllllll}
 \hline
 & Surprise & Fear & Disgust & Happy & Sad & Anger & Neutral &Average\\ \hline
 ResNet-18 & 0.72 & 0.32 & 0.49 & 0.62 & 0.30 & 0.50 & 0.33 &0.46 \\ \hline
 ARM-1 & 0.976 & 0.20 & 0.773 & 0.917 & 0.628 & 0.823 & 0.681 &0.707 \\ \hline
 ARM-2 & 0.94 & 0.30 & 0.959 & 0.964 & 0.408 & 0.950 & 0.945 &0.776 \\ \hline
 \end{tabular}
\end{table*}}

\subsection{Common Facial Expressions in Terracotta Warriors with Different Postures}
In addition to the analysis of general Terracotta Warriors, we also analyzed the expressions of Terracotta Warriors with different postures (totally 85 pictures): kneeling soldiers, Generals, vertical soldiers, cavalries and civil servant. The major expression distribution of Terracotta Warriors with different postures are as follows: for the kneeling soldiers, Disgust (38.9\%) and Sad (33.3\%); for the Generals, Disgust (68\%) and Sad (16\%); for the vertical soldiers, Surprise (35.3\%), Neutral (29.4\%); for the Cavalries, Disgust (83.3\%); for the civil servants, Fear (26.3\%), Anger (26.3\%) and Neutral(21.1\%). From a comprehensive point of view, the facial expressions of the Terracotta Warriors in different postures are mainly recognized as Happy, Sad, Anger and Neutral, combined with the historical background and position at that time, the reason why these Terracotta Warriors in different postures show such expression distribution may be that, Terracotta Warriors of Generals and soldiers usually show sadness or serious expressions in war, so they are more likely to be identified as sad and disgust expressions.

\section{Datasets and Methods}
\subsection{Collecting Pictures of Terracotta Warriors Online}
\begin{figure*}[h]
 \centering
 \includegraphics[width=0.8\linewidth]{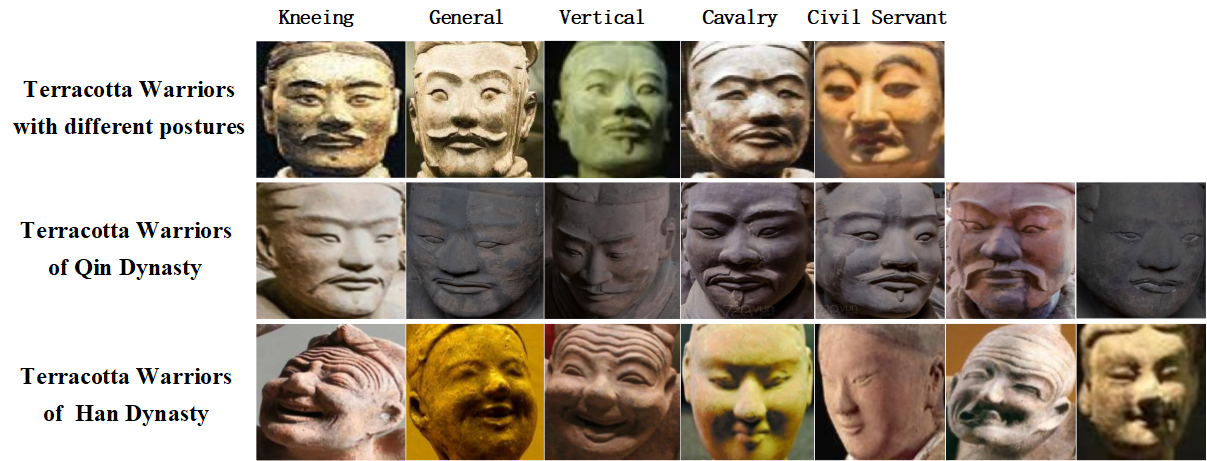}
 \caption{Some samples of Terracotta Warriors images}
 \label{fig:bmyimages}
\end{figure*}

In view of the lack of Terracotta Warriors datasets, firstly, we collect high-quality Terracotta Warriors images from Internet and VR video (from the Emperor Qinshihuang's Mausoleum Site Museum~\cite{24}, some samples of Terracotta Warriors can be seen in Figure \ref{fig:bmyimages}. We have obtained a total of 58 images of Terracotta Warriors of Qin Dynasty, 26 from Internet and 32 from VR video(we call this dataset TW58). Secondly, we collect Terracotta Warriors with different postures from Internet, which contains 5 postures, a total of 85 images: 18 kneeling, 25 
Generals, 17 vertical, 6 cavalry, 19 civil servants. Finally, we collect Terracotta Warriors images of Han Dynasty from Internet, which contains 43 images.
It is worth noting that these collected pictures are high quality ones with 68 facial landmarks detected.

\subsection{RAFDB Facial Expression Dataset}
RAFDB (13) is a large-scale affective face database with large diversities and rich annotations in real world, which includes two different subsets: single-label subset for 7 classes of basic expressions, two-tab subset for 12 classes of compound expressions. In our experiment, we use 7 basic expressions, the number of different expressions is: 1619 Surprise, 355 Fear, 877 Disgust, 5957 Happy, 2460 Sad, 867 Anger, 3204 Neutral. Some sample facial expressions of RAFDB are given in Figure \ref{fig:dataset}.

\subsection{Generating Terracotta Warriors Data by GANs}

\begin{figure}[h]
 \centering
 \includegraphics[width=0.65\linewidth]{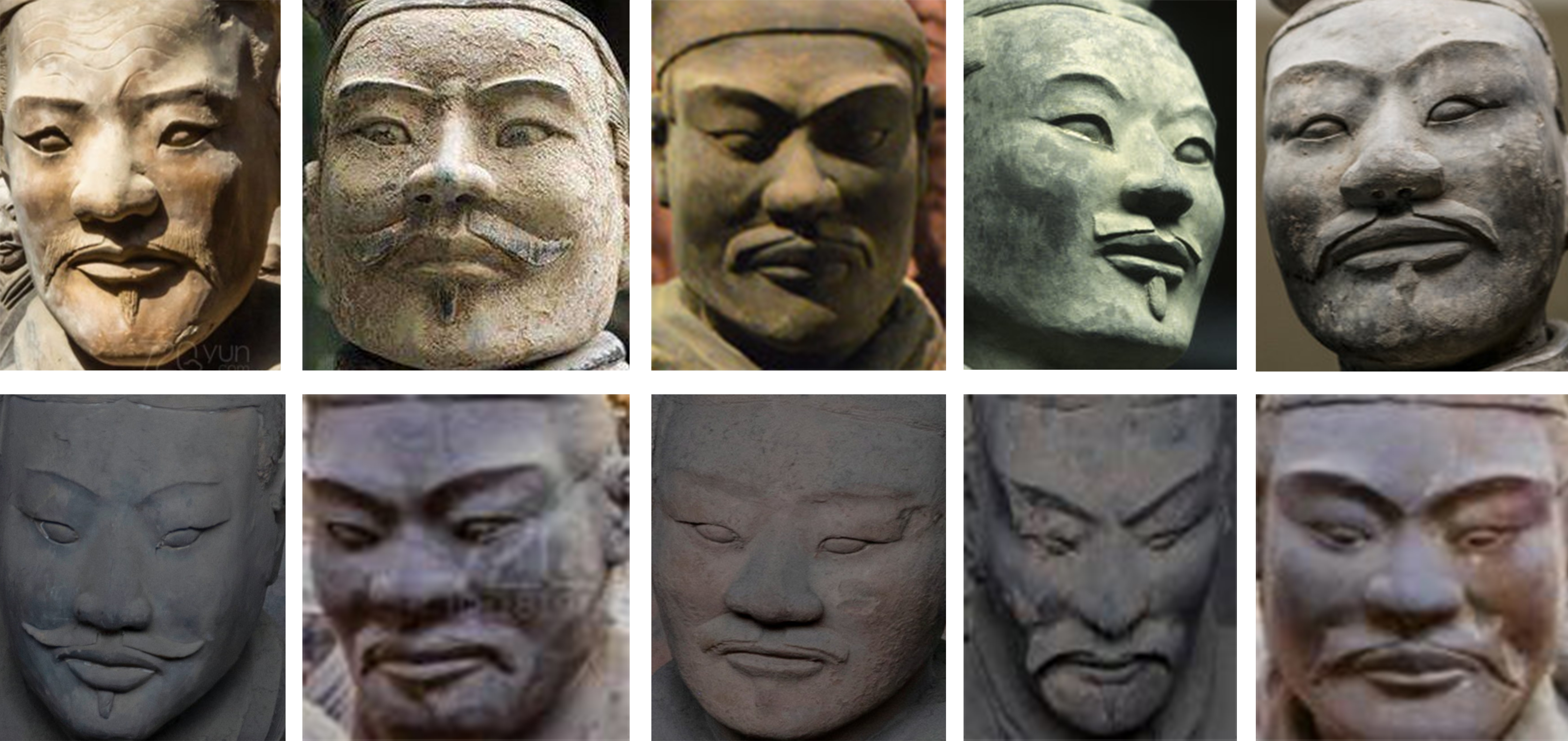}
 \caption{Some Original Terracotta Warriors Figures}
 \label{fig:ori_face}
\end{figure}

According to our research, we find that expressions of Terracotta Warriors are difficult to distinguish, as shown in Figure \ref{fig:ori_face}, it seems that every warriors have the same expression, this is a very big challenging for this research. 

\begin{figure*}
 \centering
 \includegraphics[width=0.75\linewidth]{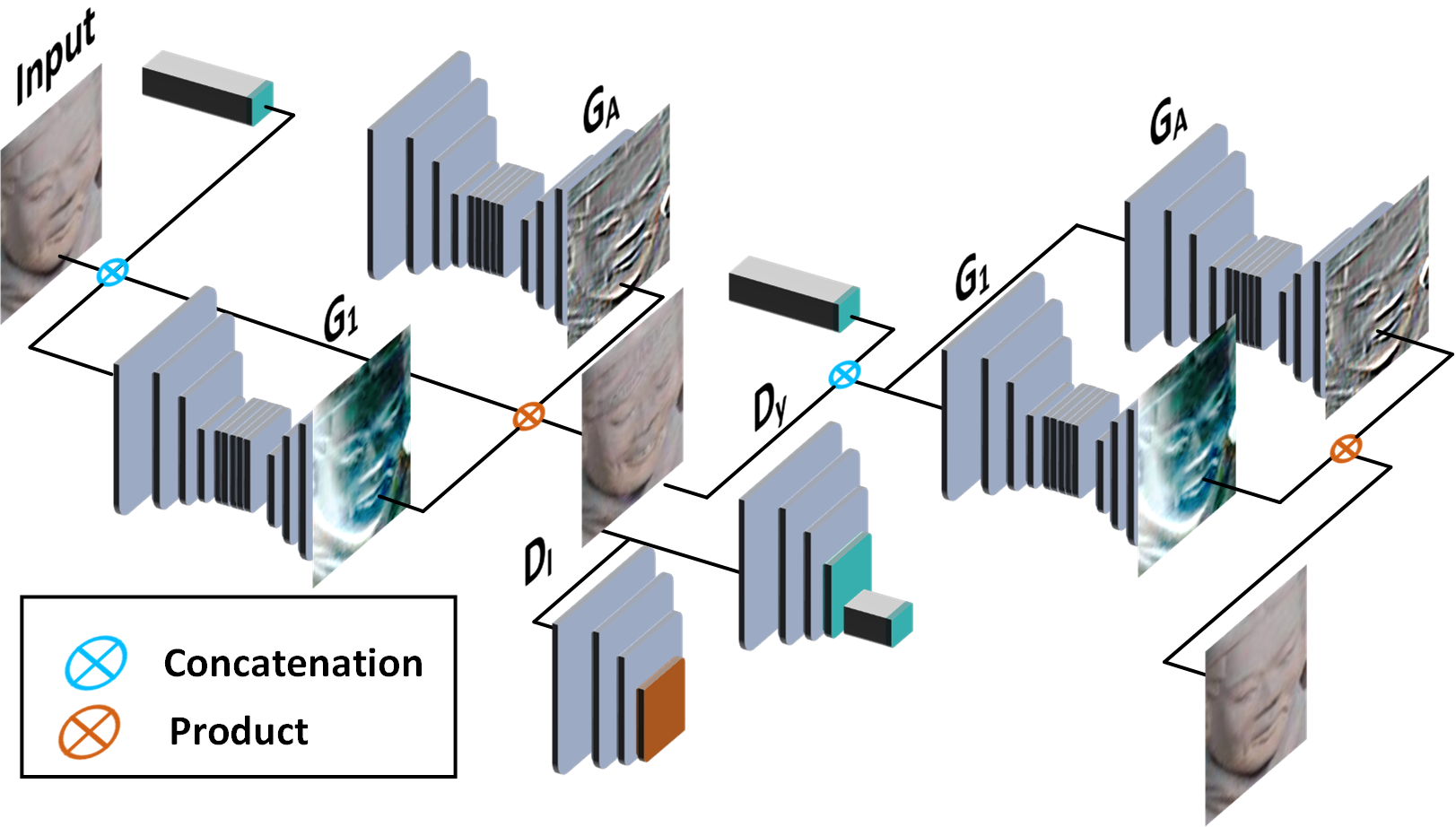}
 \caption{The structure of the GANImation we used in this paper to transfer facial expressions to Terracotta Warriors. The architecture consists of two main modules: a generator G to return attention and coloring mask; a discriminator D to evaluate the photo authenticity of the generated image.}
 \label{fig:GANImation Structure}
\end{figure*}

In order to make clear and accurate expression label for each Terracotta Warrior, we adopt a GAN network named GANimation~\cite{25} to transfer different facial expressions to Terracotta Warriors, the network's structure is shown in Figure \ref{fig:GANImation Structure}. By doing so, we can obtain Terracotta Warriors figures with much more obvious facial expressions. Then we can utilize these figures to train our expression classification network.
 
\begin{figure}
 \centering
 \includegraphics[width=0.85\linewidth]{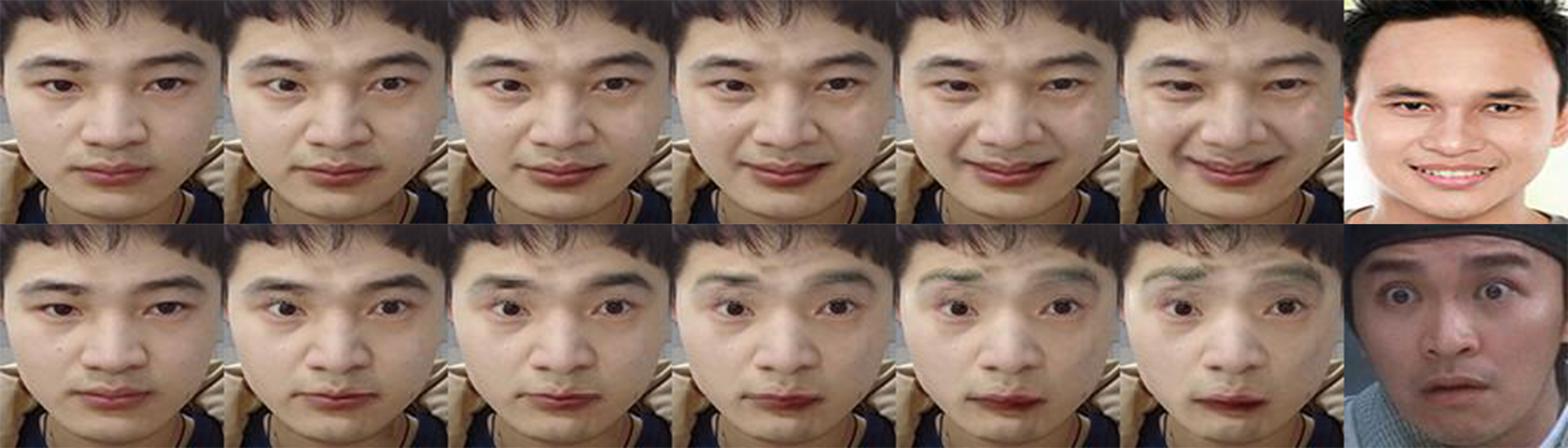}
 \caption{Some facial expressions generated by GANImation, from left to right, the generated expression is closer to the rightmost reference expression.}
 \label{fig:ganimation_sample}
\end{figure}

GANImation is an Action Units (AU) based facial expression generation network, which allows controlling the magnitude of activation of each AU and combines several of them. GANImation is not constrained to a discrete number of expressions and can animate a given image and render novel expressions in a continuum, as shown in Figure \ref{fig:ganimation_sample}.

In the first step, we must evaluate the effectiveness of GANImation, to make sure that the facial expression generated by GANImation will not affect the results of classification by other facial expression classification network. To do so, we adopt a well-known dataset called RAFDB with large-scale affective face database with large diversities and rich annotations in real world. 

\begin{figure}
 \centering
 \includegraphics[width=0.85\linewidth]{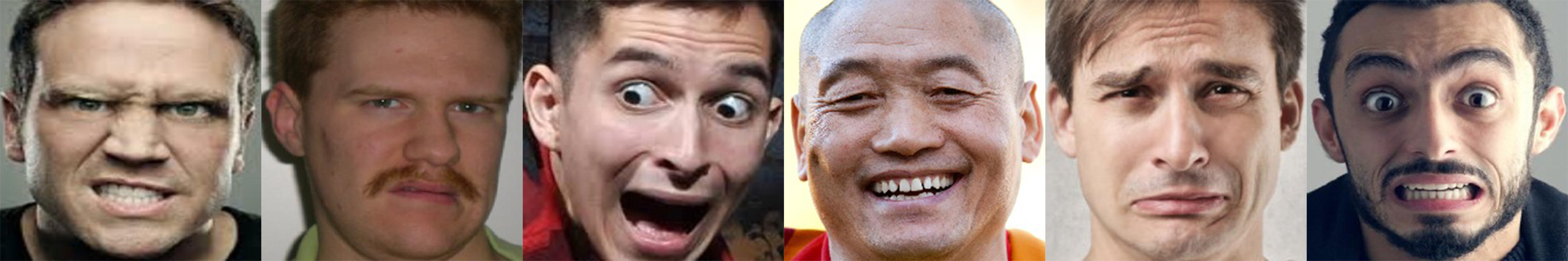}
 \caption{Some emotion figures from RAFDB,are used in GANImation to transfer expressions to figures with Neutral expression.}
 \label{fig:target_emotions}
\end{figure}

In this experiment we only use figures with Neutral expression from 7 basic expressions set. For each Neutral expression figures, we use GANImation to transfer Angry, Disgust, Fear, Happy, Sad and Surprise expression to them, the target expression figures are shown in Figure \ref{fig:target_emotions}. As a result, each Neutral expression figure will generate another six figures that has Angry, Disgust, Fear, Happy, Sad and Surprise expressions, respectively, then a new dataset is formed.

\begin{table*}
\centering
\caption{Performance of a basic facial expression recognition classifier (ResNet-18) on original RAFBD dataset and images generated by the GANImation network.}
\label{tab:GANImation_Evaluation}
\begin{tabular}{ccccccccc}
\hline
Dataset & Surprise & Fear & Disgust & Happy & Sad & Angry & Neutral & Avg\\
\hline
RAFDB & 0.93 & 0.67 & 0.78 & 0.87 & 0.86 & 0.74 & 0.77 & 0.8028 \\
GANImation & 0.89 & 0.61 & 0.72 & 0.91 & 0.88 & 0.71 & 0.77 & 0.7842 \\
\hline
\end{tabular}
\end{table*}

If the facial expression figures generated by GANImation is effective, an expression classification network should have similar performance on RAFDB dataset and dataset generated by GANImation. To evaluate this, we use a pre-trained ResNet-18 to classify facial expressions in these two datasets, and its performance is shown in Table \ref{tab:GANImation_Evaluation}, where we can observe that, on the two datasets, the prediction results of each expression are very close. Furthermore, we find that the emotion reference image is one of the main factors that affect the classification of facial expression images generated by the GANImation. In order to improve the accuracy of facial expression classification, more explicit expression reference images can be used. In conclusion, the facial expression figures generated by GANImation can achieve the same effect of expression classification as RAFDB dataset.

\begin{figure}
 \centering
 \includegraphics[width=0.85\linewidth]{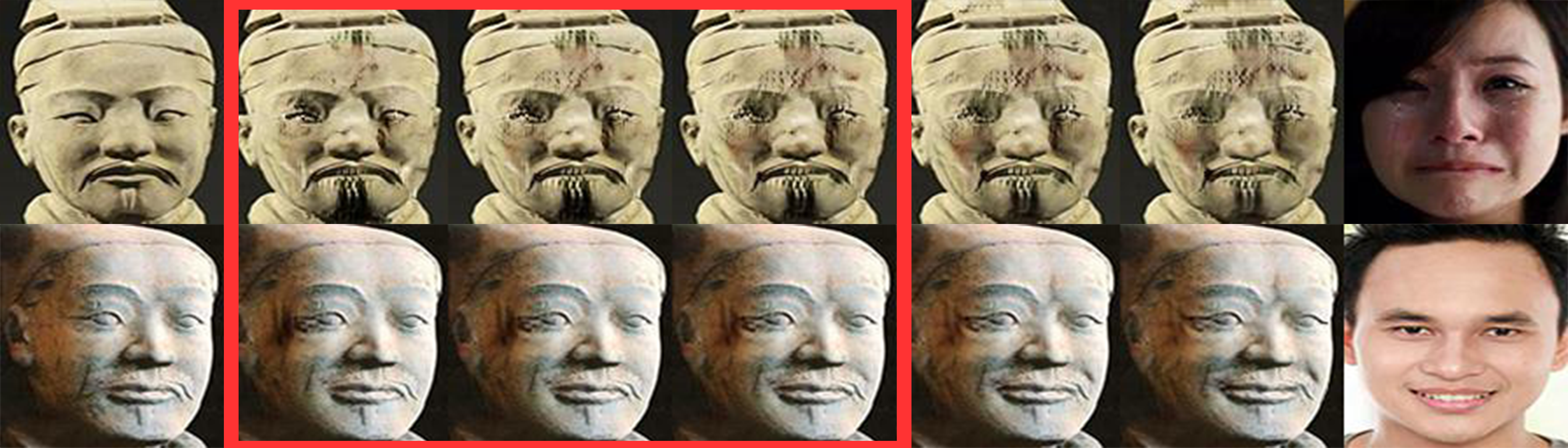}
 \caption{Terracotta Warriors facial expression generated by GANImation. From left to right, the generated expression is closer to the rightmost reference expression.}
 \label{fig:generated_emotions}
\end{figure}

After verifying the effectiveness of generated facial expression figures by GANImation, we decide to utilize GANImation network to generate 7 basic expressions for every Terracotta Warrior in TW58, some examples are shown in Figure \ref{fig:generated_emotions}. As we can observe that, there is a big gap between the expressions of Terracotta Warriors and today's real human beings. Thus, we cannot utilize the final column of the generated figures, instead, we only take figures in the red rectangular as our dataset. In total, we collect different Terracotta Warriors facial figures that we can detect Action Units from, and we generate 7 basic expressions for each of them. As a result, we can obtain 18690 Terracotta Warriors expression images. To make our dataset strong, we collect 2670 sets of 7 basic expression figures from RAFDB image dataset. Thus, we can finally have 18690 images in total, and 2690 images for each expression.

\begin{figure*}[t]
 \centering
 \includegraphics[width=0.85\textwidth]{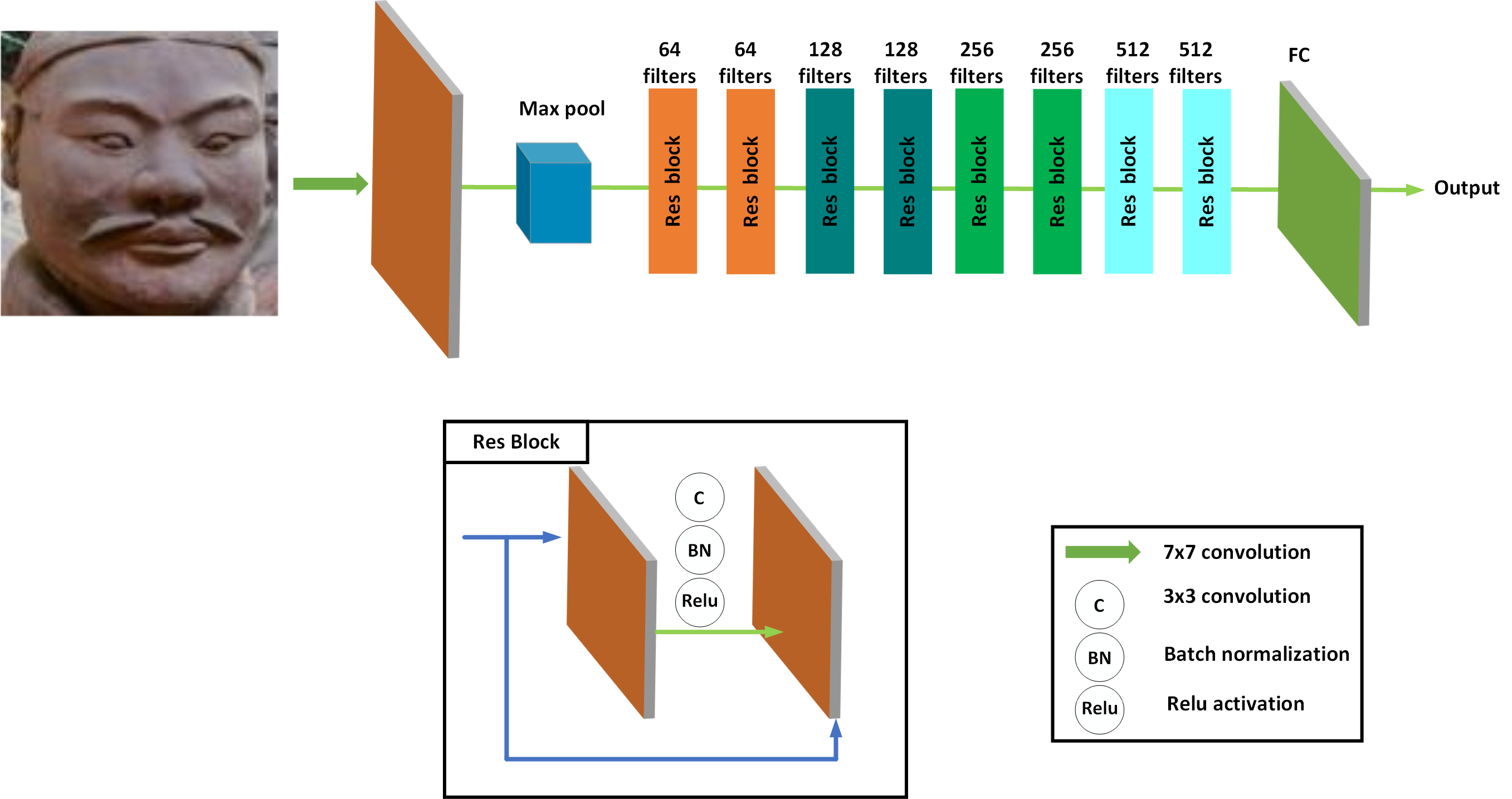}
 \caption{The basic architecture of ResNet-18 where the FC is fully connected layer. }
 \label{fig:basic architecture of ResNet}
\end{figure*}


\subsection{ResNet18 and ARM Models}
\begin{figure*}[t]
 \centering
 \includegraphics[width=0.95\textwidth]{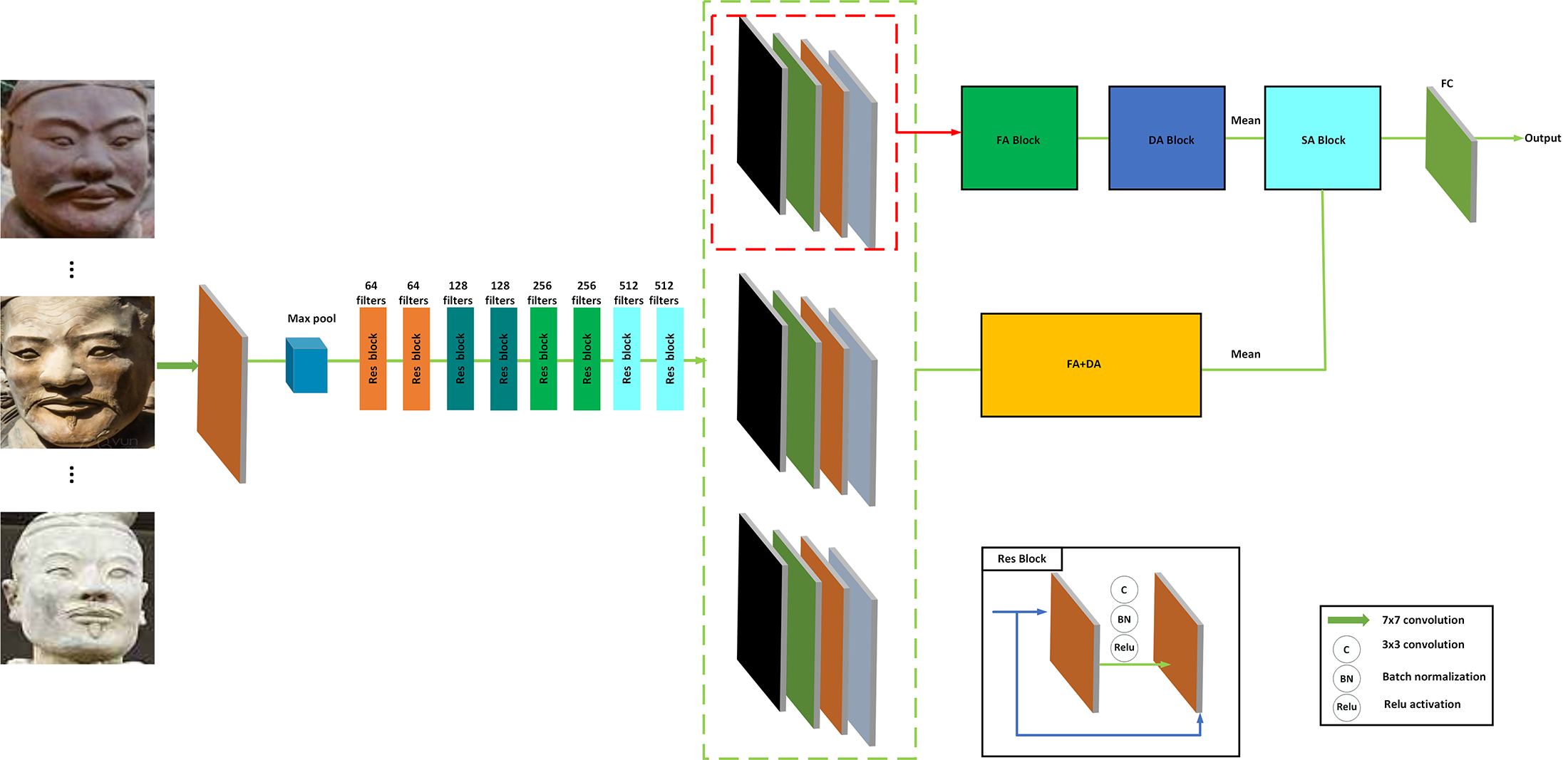}
 \caption{The basic architecture of ARM which is composed of three blocks: feature arrangement (FA) block, de-albino (DA) block, Sharing Affinity (SA) block. }
 \label{fig:basic architecture of ARM}
\end{figure*}

We also use ResNet-18~~\cite{26} as the backbone network, which is one of the most competitive network structures in current deep learning and performs well in many large-scale image recognition tasks. The basic architecture of ResNet-18 is shown in Figure \ref{fig:basic architecture of ResNet}. 
In order to make a better analysis of the Terracotta Warriors datasets we collected, we used ARM~[20] model, which has up-to-date best results for RAFDB dataset, the basic architecture of ARM can be seen in Figure \ref{fig:basic architecture of ARM}. The ARM model has a substitute for the pooling layer, theoretically it can be embedded in any convolutional neural network with a pooling layer.\\
ARM outputs high-quality representations, consists of three crucial blocks, namely Feature Arrangement (FA) block, De-albino (DA) block, and Sharing Affinity (SA) block, it should be noted that, the backbone architecture of ARM is ResNet-18, its mathematical expression is shown in formula (\ref{Eq:ResNet-18 operation1}). 
\begin{equation}
\label{Eq:ResNet-18 operation1}
\mathbf{y}=\mathcal{F}\left(\mathbf{x},\left\{W_{i}\right\}\right)+\mathbf{x}
\end{equation}
where $x$ and $y$ are the input and output vectors of the layers considered. The function ${F}(\mathbf{x},{W_{i}})$ represents the residual mapping to be learned in ResNet-18 network.

The input images of ARM model are 3-channel's colorful images with 224$\times$224 resolution, and the output are classifications of facial expressions.

In our experiments, we use the Adam Optimizer with learning rate decay strategy to train the model, Relu is used as the activation function following all convolutional layers except for the last fully connected layer. The batch size is 32, and the learning rate is set to 1e-4, all experiments are conducted using Pytorch (version 1.7) and on an Ubuntu 18.04 LTS workstation operation system with a 3.70GHz i7-8700K CPU and a V100 GPU.

\subsubsection{Pre-processing}
\subsubsection{Face Detect} In our experiment, in order to avoid the adverse effect of the background of image data on network training, and reduce training time, we use the face detector in OpenCV library to detect all Terracotta Warrior's faces, and use those face images to create a new dataset, where all images only retain the face area.

\subsubsection{Loss Function}
For training our model more effectively, and reducing overfitting condition, we adopt focal loss~\cite{27} as our loss function, that is proposed to solve the problem of sample imbalance, which can reduce the weight of easily classifying samples, and make the model focus on difficult classifying samples in training. The cross entropy~(CE) loss function can be defined as:

\begin{align}
\label{Eq:loss operation1}
CE\left( p_{\mathrm{t}} \right)=-\log \left(p_{\mathrm{t}}\right)
\end{align}

where $p_{\mathrm{t}}$ is the probability that the sample belongs to true class. 
The Focal Loss~(FL) function is modified on the basis of standard cross entropy loss, its mathematical formula can be expressed as follows: 

\begin{equation}
\mathrm{FL}\left(p_{\mathrm{t}}\right)=-\alpha_{\mathrm{t}}\left(1-p_{\mathrm{t}}\right)^{\gamma} \log \left(p_{\mathrm{t}}\right)
\end{equation}

where $\alpha_{\mathrm{t}}$ is the control parameter used to control the contribution of positive and negative samples to the total loss. It is worth noting that in the original paper~\cite{27}, the experimental results of $\gamma$ = 2 and $\alpha$ = 0.25 are the best.


\section{Discussion}

 There are some challenges and limitations in studying facial expressions of Terracotta Warriors: the facial expressions of different Terracotta Warriors are very close to each other and difficulty to distinguish, this puts a big challenge to this research; there is a lack of enough data for each category of facial expressions; the facial expressions in Terracotta Warriors are imbalance and limited, there are more numbers in Neutral, Anger and Sad than other types; there are very big facial expression differences between Terracotta Warriors (ancient art work) and today's real people which most of well-known facial expression datasets such as RAFDB and AffectNet are based on. \\

 We also apply deep transfer learning to Terracotta Warriors through RAFDB but find that the average accuracy of the classification is below 30\%. We apply trained ARM model to test the 33 American ancient arts [22] with 68 facial landmarks detected, and find major expressions are Sad, Fear and Anger, close to the results as reported in [22]. \\

 
 We compute the similarity scores of  TW58 dataset  and randomly selected 60 images from RAFDB, and find that the average similarity between TW58 dataset are around 0.54, much higher than the average similarity of 0.28 between RAFDB. These higher similarities also bring very big challenge to the classification of different expression categories for TW58 and GAN18K datasets. 
 
 Our results suggest that it is possible to use GAN-generated data to train and learn the common facial expressions of Terracotta Warriors, the proposed approach may also be applied to analyze other ancient humanistic arts. In the future, we want to explore more approaches to improve both accuracy and efficiency.\\


%



\ifCLASSOPTIONcompsoc
  \section*{Acknowledgments}
  \textbf{Funding}: This work is partially supported by National Key Research and Development Program of China with ID 2018AAA0103203.
  \\
  \textbf{Author contributions}: W.Tian  conceived main ideas and supervised the work, Y.Xie developed the formulations and performed experiments together with H.Zhang and T. Ma, Y.Xie, H.Zhang and T.Ma contribute equally, all authors discussed the results and contributed to the final manuscripts.\\
\else
  \section*{Acknowledgment}
\fi

\section*{Competing interests}
The authors declare that they have no competing interests. 

\section*{Data availability}
The data that support the findings of this study are available within the paper and the Supplementary Materials. Additional data related to this paper are available from the corresponding authors upon reasonable request. 

\section*{Code availability}

The code that supports the findings of this study are available from the corresponding author upon reasonable request.

\section*{Supplementary Materials}
        
Supplementary material for this article is available. 

\ifCLASSOPTIONcaptionsoff
  \newpage
\fi

\end{document}